\documentclass{article}
\usepackage{ijcai26}

\usepackage{times}
\usepackage{soul}
\usepackage{url}
\usepackage[hidelinks]{hyperref}
\usepackage[utf8]{inputenc}
\usepackage[small]{caption}
\usepackage{graphicx}
\usepackage{amsmath}
\usepackage{amsthm}
\usepackage{amssymb}
\usepackage{booktabs}
\usepackage{algorithm}
\usepackage{algorithmic}
\usepackage[switch]{lineno}
\usepackage{pifont}
\newcommand{\cmark}{\ding{51}}
\newcommand{\xmark}{\ding{55}}

\title{PATE-Forensics: Perception-as-Tool for Explainable Deepfake Forensics with General-Purpose MLLMs}

\author{
Yaqi Li$^1$\thanks{Equal contribution.}
\and
Jielun Peng$^1$\footnotemark[1]
\and
Yabin Wang$^1$\thanks{Corresponding author.}\and
Jincheng Liu$^1$\And
Xiaopeng Hong$^{1}$\\
\affiliations
$^1$Harbin Institute of Technology\\
\emails
\{25s103223, 25s003052\}@stu.hit.edu.cn,
wang-yabin@outlook.com,
25b903114@stu.hit.edu.cn,
hongxiaopeng@ieee.org
}

\begin{document}

\maketitle

\begin{abstract}
   Existing explainable deepfake forensic methods typically rely on task-adapted MLLM to jointly address detection, localization, and explanation. Inspired by agent-style tool use, we instead introduce a Perception-as-Tool paradigm and instantiate it as PATE-Forensics, which architecturally decouples detection and localization from explanation generation while coupling detection and localization as tightly as possible within a forensic perception tool. The DINOv3-based tool couples a multi-granularity detection module that integrates global, patch-level, and segment-level evidence with a cue-guided localization module by spatializing the patch-level and segment-level evidence into forgery score maps that guide dense mask prediction. The original image and forensic perception outputs produced by the tool form structured forensic context for a general-purpose MLLM, which is guided by prompt constraints to generate explanations without task-specific fine-tuning.  On DDL-X Track 3, PATE-Forensics achieves the best official score of 0.89, outperforming the second-ranked team by 0.19 points. Our code is available at \url{https://github.com/yqli00000/PATE-Forensics}.
\end{abstract}  

\section{Introduction}

Recent advances in deepfake generative models have significantly improved the realism and controllability of manipulated facial images~\cite{Karras_2019_CVPR,Rombach_2022_CVPR,NEURIPS2022_ec795aea,WANG2026113395}, raising serious concerns about digital security, identity trustworthiness, and visual authenticity~\cite{nguyen2022deepfakeSurvey,Peng_2026_CVPR,hopf2026robustdeepfake}. A practical forensic system should not only determine whether an image is fake, but also localize the manipulated regions and explain the supporting visual evidence in a human-readable form.


\begin{figure}[t]
    \centering
    \includegraphics[width=0.50\textwidth]{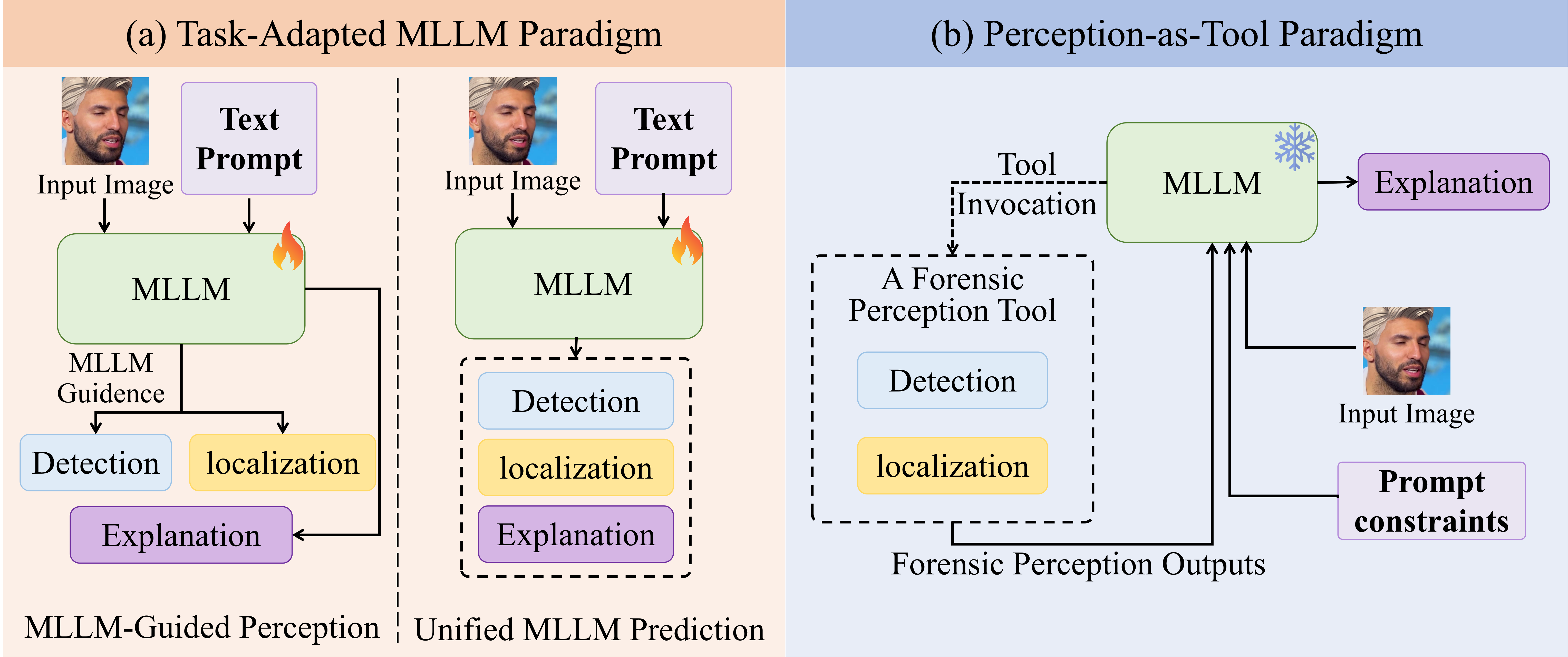}
    \caption{Comparison with existing method jointly addressing detection, localization, and explanation. Existing methods rely on task-adapted MLLMs, either using MLLM outputs to guide dedicated detection and localization modules or directly producing all three task outputs. In contrast, PATE-Forensics keeps a general-purpose MLLM frozen and externalizes detection and localization as a forensic perception tool, whose outputs, together with the original image, form the structured forensic context for explanation generation.}
    \label{fig:comparison}
\end{figure}
Motivated by these requirements, existing methods have begun to integrate forgery detection, manipulation localization, and textual explanation with the help of multimodal large language models (MLLMs)~\cite{kang2025legion}. Existing approaches mainly follow two task-adapted designs. Some methods fine-tune MLLMs on task-specific forensic data and use the generated descriptions or hidden representations to guide dedicated detection and localization modules~\cite{xu2025fakeshield,Huang_2025_CVPR,kang2025legion}. A recent unified variant instead applies reinforcement-learning-based fine-tuning to a unified MLLM, enabling it to jointly produce detection, localization, and explanation outputs~\cite{Li_2026_CVPR}. Despite their structural differences, both designs internalize domain-specific forensic capabilities through task-specific MLLM adaptation. This reliance on task-specific MLLM fine-tuning can be costly and less scalable, since each new forensic task or setting may require another round of MLLM fine-tuning. Meanwhile, fine-grained forensic perception still depends on dense visual cues such as local artifacts, texture inconsistencies, facial-part anomalies, and boundary blending patterns. 

As general-purpose vision-language models become increasingly capable in reasoning and language generation, this motivates us to investigate whether a general-purpose MLLM can generate domain-specific forensic explanations without task-specific fine-tuning when supplied with reliable structured forensic context. We therefore reorganize how the three tasks collaborate by architecturally decoupling detection and localization from explanation generation, while coupling detection and localization as tightly as possible to construct such context. Inspired by agent-style tool use, we introduce \textbf{a Perception-as-Tool paradigm}, in which coupled detection and localization are externalized as a forensic perception tool, while explanation generation is handled by a general-purpose MLLM. The tool produces a fake/real probability and a manipulation localization result. Together with the original image, these outputs form the structured forensic context used by a general-purpose MLLM to generate human-readable explanations. This design confines domain-specific visual learning to the forensic perception tool and avoids task-specific MLLM fine-tuning.

To build the reliable forensic perception tool, we further consider  how detection and localization can be coupled as tightly as possible. The key is to learn shared forensic evidence that is both discriminative for image-level detection and spatially informative for localization. Existing detectors have exploited global representations and multi-scale patch features to capture forensic evidence at different granularities~\cite{zhao2021multi,wang2022m2tr}. Nevertheless, patch-wise evidence alone may be insufficient, as facial forgery artifacts are often associated with semantically meaningful facial components, such as the eyes and mouth~\cite{Schwarcz_2021_CVPR,Haliassos_2021_CVPR}. Since a facial component generally spans multiple image patches, patch-wise modeling may fragment its forensic evidence and fail to capture consistency across the component. Segment-level evidence is therefore also needed to aggregate semantically related patches and model component-level consistency beyond individual patch responses. Meanwhile, prior manipulation localization methods show that dense mask prediction benefits from explicit forensic cues such as noise-sensitive fingerprints, or noise-guided amplification~\cite{Chen_2021_ICCV,Guillaro_2023_CVPR,cai2026zooming}. However, these cues are typically constructed specifically for localization, leaving them disconnected from the evidence learned for detection. We instead connect the two tasks through shared patch-level and segment-level forensic evidence, which supports image-level detection and is spatialized into forgery score maps to guide dense mask prediction. This design tightly couples multi-granularity detection with cue-guided localization within a single forensic perception tool.


Based on this design, we instantiate this paradigm as \textbf{PATE-Forensics} (\textbf{P}erception-\textbf{a}s-\textbf{T}ool for \textbf{E}xplainable Deepfake \textbf{Forensics}). PATE-Forensics implements a DINOv3-based forensic perception tool that couples a multi-granularity detection module and a cue-guided localization module. The detection module mines multi-granularity evidence from global, patch-level, and segment-level representations. At the global level, the DINOv3 class-token representation captures holistic image context. At the patch level, dense patch tokens are used to estimate local suspiciousness. At the segment level, dense features are grouped into semantic-aware regions to model regional consistency and estimate region-level suspiciousness. The patch-level and segment-level suspiciousness scores contribute to image-level detection and are also spatialized into score maps. The localization module takes these maps as coarse forgery cues and fuses them with multi-layer DINOv3 features through a lightweight decoder to predict a dense manipulation mask. Finally, the original image and forensic perception outputs, including the predicted probability, localization overlay, and suspicious-region crops, form the structured forensic context supplied to a general-purpose MLLM for explanation generation without task-specific fine-tuning. Official evaluation on DDL-X Track 3 shows that PATE-Forensics \textbf{ranked first} with a score of 0.89 under the official evaluation metric, outperforming the second-ranked team by 0.19 points.

Our contributions are summarized as follows:
\begin{itemize}
    \item We introduce a Perception-as-Tool paradigm and instantiate it as PATE-Forensics, which architecturally decouples detection and localization from explanation generation, and enables a general-purpose MLLM to generate forensic explanations from structured forensic context without task-specific fine-tuning.
    \item We develop a coupled forensic perception tool that combines multi-granularity detection across global, patch, and segment levels with cue-guided localization, where detection-stage patch and segment suspiciousness maps guide dense mask prediction. 
    \item Our approach ranked first in the DDL-X Challenge, demonstrating competitive performance in deepfake detection, localization, and explainability.
\end{itemize}
\section{Related Work}
\subsection{Image Forgery Detection and Localization}
Image forgery detection and localization have been studied as closely related forensic tasks. Prior studies suggest that reliable forgery analysis requires sensitivity to subtle forensic details rather than only high-level semantics. MVSS-Net~\cite{Chen_2021_ICCV} improves generalization by combining multi-view feature learning and multi-scale supervision, showing that image-level detection and pixel-level localization benefit from low-level forensic evidence. Recent methods combine heterogeneous forensic evidence more explicitly. TruFor~\cite{Guillaro_2023_CVPR} uses RGB content and a learned noise-sensitive fingerprint to produce a localization map, an image-level integrity score, and a reliability map. OpenSDI~\cite{Wang_2025_CVPR} studies diffusion-generated image spotting in open-world settings and includes both detection and localization for globally and locally manipulated images. NFA-ViT~\cite{cai2026zooming} addresses localized AIGC editing by using noise-guided attention to amplify subtle forgery cues and improve fine-grained localization. Although these methods demonstrate the importance of detailed forensic evidence, the relationship between detection and localization is usually established through shared representations, multi-task supervision, or image-level scores derived from localization outputs. Such designs couple the two tasks, but suspicious-region cues mined during detection are not explicitly reused as coarse forgery cues to guide dense mask prediction. 

For face forgery analysis, many works show that manipulation traces are often local and related to facial parts or blending boundaries. Face X-Ray~\cite{Li_2020_CVPR} detects face forgeries by revealing blending boundaries shared by many face manipulation pipelines. Lips Don't Lie~\cite{Haliassos_2021_CVPR} focuses on mouth dynamics for generalizable face forgery detection, while parts-based detectors~\cite{Schwarcz_2021_CVPR} analyze artifacts around individual facial components. More recently, Han et al.~\cite{Han_2025_CVPR} introduce facial component guidance to adapt a foundation model for video-based deepfake detection, encouraging the model to attend to key facial regions for improved generalization. These works suggest that local artifacts, facial component consistency, and region-level structures are all important for robust forgery analysis.

Our method follows this motivation but differs in how local and regional evidence are constructed and used. Instead of relying only on a fused dense representation or manually defined facial parts, we exploit DINOv3 dense features~\cite{cuttano2026insid3} to build patch and segment-level evidence. The patch branch captures patch-level evidence, while the segment branch models suspiciousness over spatially coherent regions. Both branches contribute to image-level detection, and their suspiciousness scores are further converted into score maps to guide dense mask prediction, forming a coupled detection-localization design within a single forensic perception tool.

\subsection{Explainable Image Forgery Analysis}

In forensic applications, a binary prediction alone is often insufficient. Users need to know where the suspicious evidence lies and why the system regards an image as manipulated. Recent multimodal methods have therefore moved toward jointly addressing detection, localization, and explanation through task-adapted MLLMs. These approaches can be organized into two designs. The first is MLLM-guided perception, where MLLM outputs or representations assist dedicated perception modules. FakeShield~\cite{xu2025fakeshield} generates a textual tampering description and uses its representation to guide a forgery localization module. SIDA~\cite{Huang_2025_CVPR} extracts detection and segmentation token representations from a large vision-language model and feeds them into dedicated heads for authenticity prediction and mask generation. The second is unified MLLM prediction. Omni-Fake-R1~\cite{Li_2026_CVPR} is a recent example that applies reinforcement-learning-based fine-tuning to a unified omni-modal model, enabling it to jointly produce authenticity decisions, localization results, and natural-language explanations.

Although the two designs differ in how the three tasks are connected, both adapt the MLLM to domain-specific forensic data and output requirements. Our method instead follows a Perception-as-Tool paradigm. A forensic perception tool performs detection and localization without receiving MLLM guidance, and its outputs, together with the original image, form structured forensic context for a general-purpose MLLM. Thus, the MLLM neither assists forensic perception nor directly predicts all three task outputs. It uses the structured forensic context only for interpretation and explanation generation, avoiding task-specific fine-tuning.

\section{Methodology}

\begin{figure*}[t]
    \centering
    \includegraphics[width=1\textwidth]{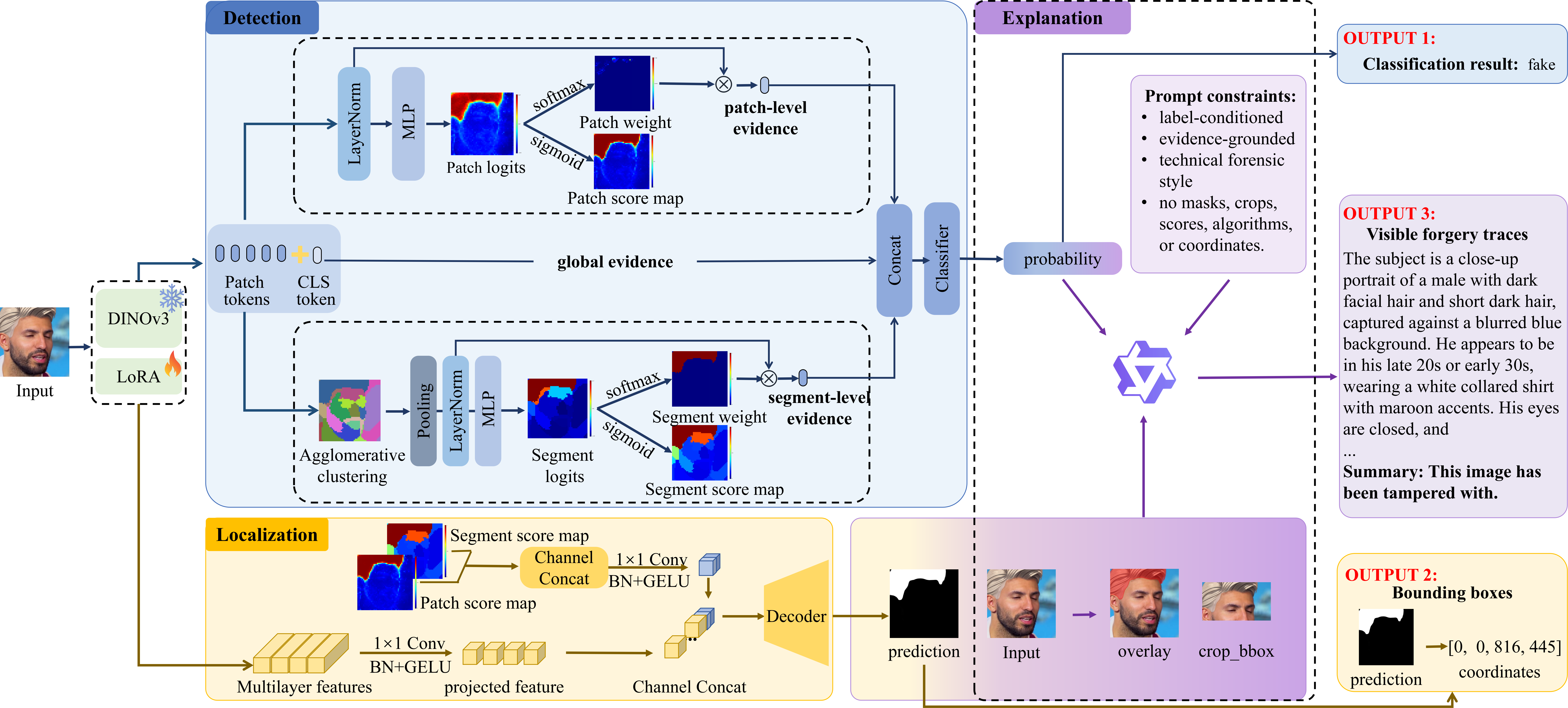}
    \caption{\textbf{Overview of PATE-Forensics.} Following the Perception-as-Tool paradigm, a forensic perception tool performs detection and localization through two coupled modules. The global, patch, and segment branches provide multi-granularity evidence for fake/real detection, while the cue-guided localization module fuses detection-stage patch and segment score maps with multi-layer DINOv3 features to predict a dense forgery mask. The original image and forensic perception outputs form structured forensic context for a general-purpose MLLM. Guided by prompt constraints, the MLLM generates the final explanation without task-specific fine-tuning.}
    \label{fig:framework}
\end{figure*}
PATE-Forensics employs a DINOv3-based forensic perception tool that couples multi-granularity detection with cue-guided localization, without relying on MLLM assistance. The original image and forensic perception outputs are organized as structured forensic context for a general-purpose MLLM to generate human-readable forensic explanations without task-specific fine-tuning. Figure~\ref{fig:framework} provides an overview of PATE-Forensics.

Given an input image $I$, the forensic perception tool extracts a class token and dense patch tokens using a DINOv3 backbone with LoRA adaptation. Its detection module integrates global, patch-level, and segment-level evidence for fake/real detection. The patch and segment branches not only contribute to image-level detection, but also produce coarse forgery cues in the form of patch and segment score maps. The localization module fuses these coarse forgery cues with multi-layer DINOv3 features through a lightweight cue-guided decoder to predict a dense forgery mask. During inference, the original image, predicted probability, dense mask, overlaid image, and bounding-box crops are organized as structured forensic context for explanation generation.

\subsection{Multi-Granularity Forgery Detection}
The detection module performs image-level fake/real prediction by integrating global, patch-level, and segment-level evidence. The global branch captures holistic image context from the DINOv3 class token. The patch branch estimates patch-level suspiciousness from dense patch tokens, while the segment branch aggregates patch tokens into semantic-aware region prototypes to model spatially coherent manipulation patterns. In this way, the detector integrates image-level context with patch-level, and segment-level evidence, where the latter captures regional consistency beyond individual patches.

\paragraph{Global evidence.}
The global evidence is derived from the DINOv3 class token. The normalized class token serves as the global representation in the final multi-granularity detector and is also passed through a lightweight classifier to produce an auxiliary image-level fake/real logit:
\begin{equation}
    \ell_g^{img} = C_g(\mathrm{LN}(z_{cls})).
\end{equation}

\paragraph{Patch-level evidence.}
The patch branch operates on the trainable DINOv3 patch tokens. Let $X=\{x_i\}_{i=1}^{N}$ denote the normalized patch tokens, where $N$ is the number of image patches. For each patch token, the patch branch predicts a patch-level suspiciousness logit:
\begin{equation}
    a_i = f_p(x_i),
\end{equation}
where $f_p$ is a lightweight MLP and $a_i$ measures how suspicious the $i$-th patch is. The resulting logits serve two roles. First, they are converted into temperature-scaled weights and used to aggregate patch tokens into an image-level patch representation $p_{agg}$, which serves as the patch-level representation in the final multi-granularity detector:
\begin{align}
    w_i &= \frac{\exp(a_i / \tau)}{\sum_{j=1}^{N}\exp(a_j / \tau)}, \\
    p_{agg} &= \sum_{i=1}^{N} w_i x_i .
\end{align}
The aggregated representation $p_{agg}$ is then passed to the patch classifier to produce an auxiliary image-level fake/real logit:
\begin{equation}
    \ell_p^{img} = C_p(\mathrm{LN}(p_{agg})).
\end{equation}
Second, the same patch-level logits are spatialized into a score map. After applying the sigmoid function and reshaping the logits to the patch grid, we obtain a patch-level coarse forgery cue that provides spatial guidance for localization: 
\begin{equation}
    S_p = \sigma(\mathrm{reshape}(a)),
\end{equation}
where $a=[a_1,\ldots,a_N]$. In this way, the patch branch produces both image-level evidence for detection and spatial cues for localization.

\paragraph{Segment-level evidence.}
The segment branch introduces region-level evidence by grouping patch tokens into spatially coherent segments. To avoid unstable assignments caused by task-specific adaptation, agglomerative clustering is performed on frozen DINOv3 patch tokens. The resulting assignments are then applied to the LoRA-adapted patch tokens. This produces a patch-to-segment assignment $r_i \in \{1,\ldots,K\}$ for each patch $i$, where $K$ is the number of semantic-aware regions. Given the trainable patch tokens $X=\{x_i\}_{i=1}^{N}$, the prototype of segment $k$ is computed by averaging the tokens assigned to that segment:
\begin{equation}
    q_k = \frac{1}{|\mathcal{R}_k|}\sum_{i \in \mathcal{R}_k} x_i,
    \quad
    \mathcal{R}_k=\{i \mid r_i=k\}.
\end{equation}
Analogously to the patch branch, each segment prototype is assigned a suspiciousness logit:
\begin{equation}
    b_k = f_s(q_k),
\end{equation}
where $f_s$ is a lightweight MLP and $b_k$ measures the suspiciousness of segment $k$. These logits are used to obtain both an image-level segment representation and a segment score map:
\begin{align}
    s_{agg} &= \sum_{k=1}^{K}
    \frac{\exp(b_k / \tau)}{\sum_{j=1}^{K}\exp(b_j / \tau)}
    q_k, \\
    \ell_s^{img} &= C_s(\mathrm{LN}(s_{agg})). \\
    S_s(i) &= \sigma(b_{r_i}).
\end{align}
Here, $s_{agg}$ serves as the segment-level representation in the final multi-granularity detector, while $S_s$ provides a spatially coherent coarse forgery cue for localization. Compared with the patch score map, $S_s$ is more region-consistent because all patches in the same segment share the same suspiciousness score.

\paragraph{Main detection.}
The final detector fuses the normalized global feature, aggregated patch feature, and aggregated segment feature:
\begin{equation}
    \ell_m = C_m([\mathrm{LN}(z_{cls}); \mathrm{LN}(p_{agg}); \mathrm{LN}(s_{agg})]),
\end{equation}
where $z_{cls}$, $p_{agg}$, and $s_{agg}$ denote the class-token representation, aggregated patch representation, and aggregated segment representation, respectively. The main classifier $C_m$ produces the final fake/real logit $\ell_m$.

\subsection{Cue-Guided Localization}
The localization module predicts a dense manipulation mask conditioned on the coarse forgery cues produced by the patch and segment branches. In parallel, patch tokens from multiple hidden layers of DINOv3 are reshaped into spatial feature maps and used as multi-level visual inputs to the decoder. The patch and segment score maps are concatenated into a two-channel guidance map:
\begin{equation}
    S = \mathrm{concat}(S_p, S_s).
\end{equation}
The decoder follows a lightweight SegFormer-style design~\cite{NEURIPS2021_64f1f27b}. Each DINOv3 feature map is projected into a shared embedding dimension by a $1 \times 1$ convolution followed by batch normalization and GELU activation. The two-channel score map is projected in the same way. All projected features are concatenated along the channel dimension. 
\begin{align}
    F_{merge} =
    \mathrm{Concat}\big(&
    \phi_1(F_1), \ldots,
    \phi_4(F_4),
    \phi_s(S)\big), \\
    \hat{M} &= \sigma(\mathrm{Up}(\psi(F_{merge}))).
\end{align}
Here $F_1,\ldots,F_4$ are selected DINOv3 feature maps, $\phi_l$ and $\phi_s$ denote projection layers for DINOv3 features and score maps, $\psi$ is the convolutional fusion head, and $\mathrm{Up}$ upsamples the mask logits to the input image resolution.

This design makes localization guided by detection-stage evidence, as the decoder uses patch and segment score maps as coarse forgery cues and fuses them with dense DINOv3 features.

\subsection{Training Objective}
The training objective supervises three groups of outputs. First, image-level fake/real labels supervise the main detector and the three auxiliary image-level logits. Let $y\in\{0,1\}$ denote the image-level fake/real label, where $y=1$ indicates a fake image. The main logit $\ell_m$, global-branch image-level logit $\ell_g^{img}$, patch-branch image-level logit $\ell_p^{img}$, and segment-branch image-level logit $\ell_s^{img}$ are all supervised by $y$:
\begin{equation}
    \begin{aligned}
    \mathcal{L}_{det}
    =&\ \mathcal{L}_{cls}(\ell_m, y)
    + \mathcal{L}_{cls}(\ell_g^{img}, y) \\
    &+ \mathcal{L}_{cls}(\ell_p^{img}, y)
    + \mathcal{L}_{cls}(\ell_s^{img}, y).
    \end{aligned}
\end{equation}

Second, the patch and segment suspiciousness logits are supervised by targets derived from the ground-truth mask. Specifically, the ground-truth mask is downsampled to patch-level targets $M_p=\{m_i^p\}_{i=1}^{N}$, and segment-level targets $M_s=\{m_k^s\}_{k=1}^{K}$ are obtained by aggregating $M_p$ within each cluster. The corresponding cue supervision loss is
\begin{equation}
    \mathcal{L}_{cues}
    =
    \mathcal{L}_{cue}^{p}(a, M_p)
    +
    \mathcal{L}_{cue}^{s}(b, M_s),
\end{equation}
where $a=[a_1,\ldots,a_N]$ are patch suspiciousness logits and $b=[b_1,\ldots,b_K]$ are segment suspiciousness logits. In the implementation, focal loss is used for these classification and mask-derived supervision terms.

Third, the final dense mask prediction is supervised with binary cross-entropy and Dice loss:
\begin{equation}
    \mathcal{L}_{mask}
    =
    \mathcal{L}_{BCE}(\hat{M}, M)
    +
    \mathcal{L}_{Dice}(\hat{M}, M),
\end{equation}
where $M$ denotes the ground-truth forgery mask. The overall training objective is defined as
\begin{equation}
    \mathcal{L}
    =
    \mathcal{L}_{det}
    +
    \mathcal{L}_{cues}
    +
    \mathcal{L}_{mask}.
\end{equation}

\subsection{Inference and Post-processing}
At inference, each image is resized to $768 \times 768$ and passed through the model. The fake probability is computed from the sigmoid of the main detection logit. Images with fake probability greater than or equal to 0.5 are classified as fake, otherwise they are classified as real.

For images predicted as fake, the predicted mask is resized back to the original image resolution and binarized with threshold 0.4. We extract 8-connected components from the binary mask and discard components with area smaller than 8 pixels. Each remaining component is converted into a bounding box $[x_1,y_1,x_2,y_2]$ in the original image coordinate system. These bounding boxes serve as the localization output for official evaluation and are also used to crop suspicious regions for explanation generation. Following the submission format, the bounding box coordinates are normalized to a 0-1000 scale:
\begin{align}
    x' = \mathrm{round}\left(\frac{x}{W} \times 1000\right) \\
    \quad
    y' = \mathrm{round}\left(\frac{y}{H} \times 1000\right),
\end{align}
where $W$ and $H$ are the original image width and height. Real images return an empty bounding-box list.

\subsection{MLLM-based Explanation Generation}
The explanation module supplies a general-purpose MLLM with structured forensic context formed by the original image and the outputs of the forensic perception tool. Explanation generation is further guided by a text prompt that specifies task and output constraints. The MLLM is used without task-specific fine-tuning. For each sample, the context includes the original image and the predicted fake probability.
When the image is predicted as fake, the localization output is further used to construct two region-level visual cues, namely an image with the predicted suspicious region highlighted and a local crop of the bounding-box region extracted from the predicted mask during inference. These inputs allow the model to inspect both the full image context and the localized region indicated by the model.

The prompt is designed with four constraints. First, it is label-conditioned. Fake samples require localized forensic evidence, while real samples require descriptions of visual consistency and the absence of manipulation artifacts. Second, it is grounded in structured forensic context. The MLLM is guided to focus on the model-indicated region and compare it with adjacent normal regions. Third, it constrains the writing style to a technical forensic annotation, using stable categories such as texture inconsistency, lighting mismatch, edge artifacts, anatomical irregularity, reflection inconsistency, and resolution or noise mismatch when applicable. Fourth, it prevents the final explanation from exposing internal implementation details such as masks, overlays, crops, thresholds, confidence scores, algorithms, or coordinates. These constraints guide the MLLM in transforming structured forensic context into coherent, well-organized, and evidence-grounded natural-language explanations. The complete prompt templates used to generate explanations for real and fake images are provided in our code repository.

\section{Experiment}
The task follows the DDL-X Track 3 setting of the IJCAI 2026 AI Safety Workshop, which jointly evaluates deepfake detection, localization, and explainability~\cite{ddlx2026track3}. We use the image portion of the Deepfake Detection and Localization Image (DDL-I) dataset~\cite{miao2025ddl}. DDL-I is a large-scale image deepfake dataset with pixel-level forgery region annotations. It consists of 1.2 million images and covers both single-face and multi-face scenarios, where manipulated regions may appear in local facial areas or in one or more faces within a multi-face image. 

\subsection{Training Data and Augmentation}
The model is trained on the DDL-I dataset. To make full use of the annotated data, we repartition the dataset into training and validation sets. Specifically, we merge the original training and validation sets and use both for training. We further randomly split the original test set into training and validation subsets at a ratio of 9:1, using a fixed random seed of 42. Consequently, the final training set consists of the original training set, the original validation set, and 90\% of the original test set, while the remaining 10\% of the original test set is used as the validation set.

Two forms of data augmentation are applied during training. First, synchronized geometric transformations are applied to each image-mask pair, including horizontal flipping and rotations by 90, 180, and 270 degrees. The training resolution is randomly sampled between 384 and 768 pixels and constrained to be divisible by the patch size. Images are resized with bilinear interpolation, while masks use nearest-neighbor interpolation to preserve binary labels.

Second, we introduce canvas-level layout augmentation to simulate the diverse layouts observed in the test set. With probability 0.25, an image-mask pair is transformed by one of three layout operations: shrinking the image and pasting it onto a solid-color canvas, repeating multiple resized copies on the canvas, or composing multiple samples into a mosaic layout. The same paste geometry is applied to the image and its mask, while the mask canvas is initialized as background. For mosaic augmentation, the image-level label is set to fake if any constituent sample is fake.

\subsection{Implementation Details}
 The model uses DINOv3-L/16 with feature dimension 1024 and patch size 16. LoRA is enabled with rank 32, scaling factor 16, and dropout 0.05. The DINOv3 feature layers selected for mask decoding are layers 6, 12, 18, and 24. The hidden dimension of the mask decoder is set to 256. Training uses AdamW with learning rate $10^{-4}$, weight decay $5 \times 10^{-2}$, mixed precision, and a warmup cosine scheduler. The model is trained for 20 epochs using distributed training on three NVIDIA RTX 3090 GPUs, with a per-GPU batch size of 8 and an effective batch size of 24. For explanation generation, we use Qwen3.5-Flash model (\texttt{qwen3.5-flash}) as the general-purpose MLLM without task-specific fine-tuning, accessed through the OpenAI-compatible API of Alibaba Cloud Model Studio (DashScope). Each request uses a maximum answer length of 700 tokens. Temperature and top-$p$ are not explicitly specified and therefore follow the provider defaults. 

\subsection{Evaluation Protocol}
Following the official DDL-X Track 3 evaluation protocol, we evaluate the system from three aspects: image-level detection, forgery localization, and explanation generation. Detection is measured by accuracy (ACC). Localization is measured by bounding-box IoU and is evaluated on fake images. Explanation quality is measured by BERTScore and a rubric-based score. The rubric-based score further evaluates factual correctness, visual grounding, regional specificity, label faithfulness, alignment between the textual explanation and predicted bounding boxes, face-centered analysis, and clarity of the generated explanation. The scores for fake and real images are calculated separately as follows:
\begin{align}
\mathrm{Score_{f}} &=
\begin{cases}
0.2A + 0.5I, & I < 0.7, \\
0.2A + 0.5I + 0.1B + 0.2R, & I \ge 0.7,
\end{cases}  \\
\mathrm{Score_{r}} &=
0.2A + 0.2B + 0.6R
\end{align}

where $A$, $I$, $B$, and $R$ denote ACC, IoU, BERTScore, and the rubric-based score, respectively. The final reported score is obtained by computing the score for each sample using the formula corresponding to its ground-truth category, i.e., real or fake, and then aggregating the resulting sample-level scores over the entire evaluation set.

\subsection{Experimental Results}

\begin{table}[t]
\centering
\small
\setlength{\tabcolsep}{3pt}
\begin{tabular}{@{}clccccc@{}}
\toprule
Rank & Team & Score & ACC & IoU & BERT & Rub. \\
\midrule
\textbf{1} & \textbf{HIT\_VIRLAB} & \textbf{0.8940} & \textbf{0.1995} & \textbf{0.3079} & \textbf{0.0943} & \textbf{0.2923} \\
2 & CBSR-WTD & 0.7025 & 0.1776 & 0.2495 & 0.0813 & 0.1941 \\
3 & Track Killer & 0.6857 & 0.1562 & 0.2363 & 0.0851 & 0.2081 \\
4 & hellohello & 0.6181 & 0.1601 & 0.2200 & 0.0769 & 0.1610 \\
5 & XQ999 & 0.5770 & 0.1601 & 0.1985 & 0.0585 & 0.1599 \\
\bottomrule
\end{tabular}
\caption{Final results compared with other teams on DDL-X Track 3. CBSR-WTD denotes CBSR\_what\_the\_dog\_doing.}
\label{tab:challenge-results}
\end{table}

Table~\ref{tab:challenge-results} reports the final DDL-X Track 3 leaderboard results. The metric columns follow the official leaderboard presentation and correspond to the weighted score components used in the final score. Our method ranked first among all participating teams, with a final score of 0.8940.

The improvement is consistent across the three evaluated aspects. Our method obtains the highest detection component and the highest localization component among the top-5 submissions, suggesting that coupling multi-granularity detection with cue-guided localization is effective for both image-level prediction and bounding-box localization. It also achieves the strongest BERTScore and rubric-based score, showing that a general-purpose MLLM guided by prompt constraints can generate effective explanations from structured forensic context comprising the original image and forensic perception outputs. These results demonstrate the practical viability of the Perception-as-Tool design without task-specific fine-tuning.

\subsection{Qualitative Analysis}
\begin{figure}[t]
    \centering    
    \includegraphics[width=\linewidth]{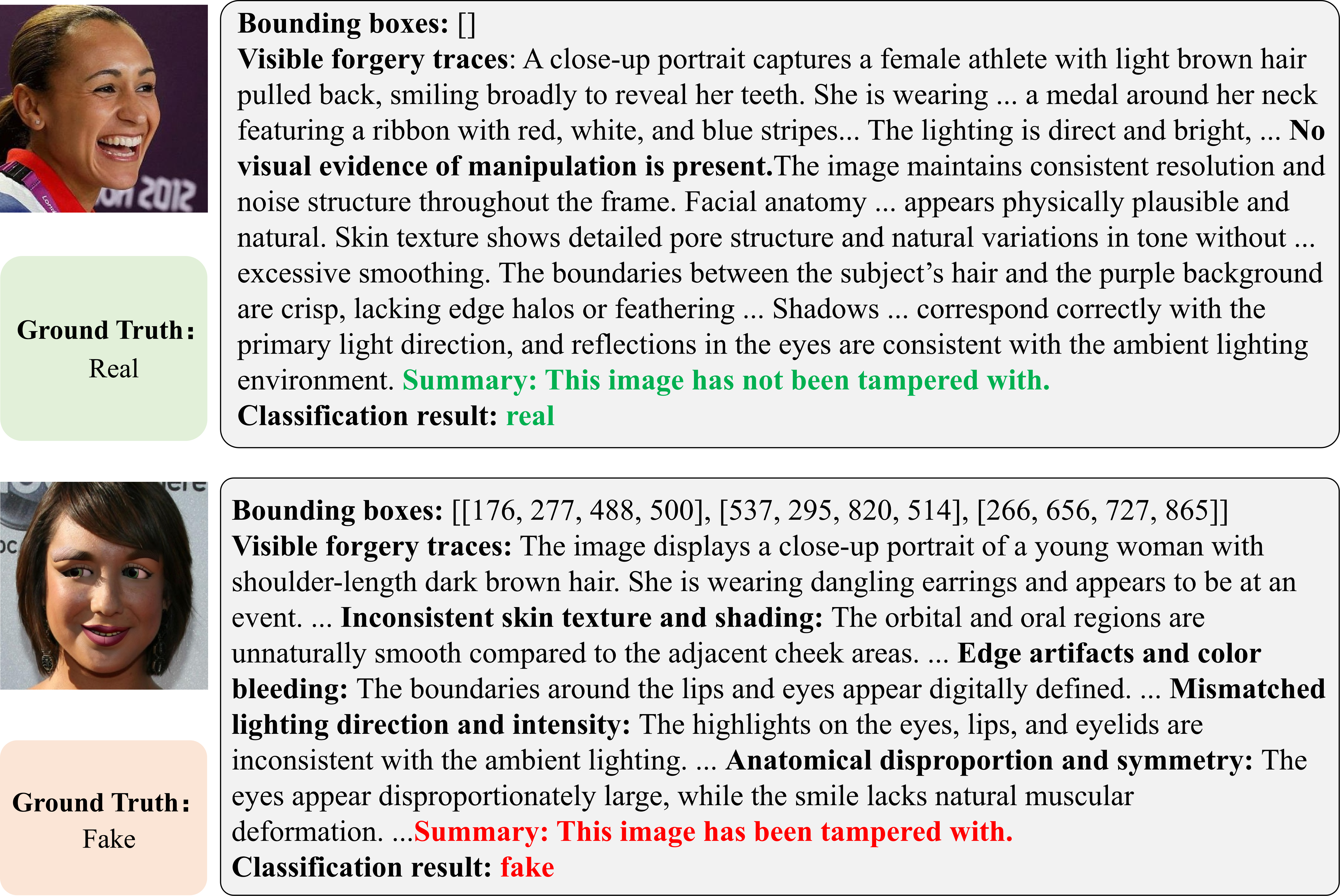}
    \caption{\textbf{Qualitative examples of MLLM-based explanation.} For the real image (top), no manipulated region is detected. For the fake image (bottom), the predicted bounding-box coordinates localize the manipulated regions. The explanations are abridged for readability.}
    \label{fig:qualitative-examples}
\end{figure}

Figure~\ref{fig:qualitative-examples} presents the explanations generated
using the structured forensic context and prompt guidance. For the real example, the perception tool predicts the image as real and identifies no suspicious region. The MLLM then supports this prediction by describing visual consistency in facial structure, texture, boundaries, illumination, and reflections. For the fake example, the perception tool predicts the image as fake and localizes candidate suspicious regions. Using the structured forensic context, the MLLM examines these regions and describes visible forensic evidence, including inconsistent skin texture and shading, boundary artifacts, color bleeding, and illumination mismatch. Rather than merely repeating the predicted label, the generated explanation connects the perception outputs with observable visual evidence. Overall, the perception
tool supplies the structured forensic context, while the prompt guides the MLLM to organize the relevant evidence into a structured explanation. This design produces label-consistent and region-aware explanations without task-spe cific MLLM fine-tuning.

\subsection{Ablation Studies}
We conduct ablation studies to examine the contributions of multi-granularity forgery detection and cue-guided localization. The training and validation sets used in all ablation experiments follow the same data-partitioning protocol described above. To ensure a fair performance evaluation, we tested all variants on the development set provided during Phase I.\footnote{\url{https://huggingface.co/datasets/zy23333/DDL-X/tree/main}} The progressive component-wise ablation results are reported in Table~\ref{tab:ablation-components}. 

Starting from the baseline, progressively adding patch-level and segment-level evidence consistently improves both ACC and IoU, indicating that the three levels of detection evidence provide complementary information. Introducing the patch and segment score maps $S_p$ and $S_s$ as coarse localization cues produces the most substantial gain, increasing IoU from 0.7410 to 0.7910. Compared with the baseline, PATE-Forensics improves ACC by 0.0047 and IoU by 0.0627, demonstrating the effectiveness of both multi-granularity forgery detection and cue-guided localization.

\begin{table}[t]
\centering
\scriptsize
\setlength{\tabcolsep}{2.0pt}
\renewcommand{\arraystretch}{1.08}
\begin{tabular}{@{}lcccccccc@{}}
\toprule
& \multicolumn{3}{c}{Detection Evidence}
& \multicolumn{3}{c}{Localization Input}
& \multicolumn{2}{c}{Metric} \\
\cmidrule(lr){2-4}\cmidrule(lr){5-7}\cmidrule(l){8-9}
Configuration
& $z_{cls}$ & $p_{agg}$ & $s_{agg}$
& $F_1$--$F_4$ & $S_p$ & $S_s$
& ACC & IoU \\
\midrule
Baseline
& \cmark & \xmark & \xmark
& \cmark & \xmark & \xmark
&  0.9914 & 0.7283 \\
$+$ Patch-level evidence
& \cmark & \cmark & \xmark
& \cmark & \xmark & \xmark
& 0.9925 & 0.7375 \\
$+$ Segment-level evidence
& \cmark & \cmark & \cmark
& \cmark & \xmark & \xmark
& 0.9939 & 0.7410 \\
$+$ Coarse forgery cues
& \cmark & \cmark & \cmark
& \cmark & \cmark & \cmark
& \textbf{0.9961} & \textbf{0.7910} \\
\bottomrule
\end{tabular}
\caption{Progressive component-wise ablation of PATE-Forensics. Components are added cumulatively from top to bottom. The baseline uses $z_{cls}$ for detection and multi-layer features $F_1$--$F_4$ for localization. The final row denotes the full PATE-Forensics model.}
\label{tab:ablation-components}
\end{table}

\section{Conclusion}
This paper introduced the Perception-as-Tool paradigm and instantiated it as PATE-Forensics for explainable deepfake forensics without task-specific MLLM fine-tuning. PATE-Forensics architecturally decouples detection and localization from explanation generation while coupling detection and localization as tightly as possible within a DINOv3-based forensic perception tool. The tool couples multi-granularity detection with cue-guided localization by spatializing patch-level and segment-level forgery logits into score maps that guide dense mask prediction. The original image and forensic perception outputs produced by the tool form the structured forensic context used by a general-purpose MLLM to generate human-readable explanations without task-specific fine-tuning. PATE-Forensics achieved the top official score on DDL-X Track 3. More broadly, our results suggest that domain perception can be externalized as a reliable tool, allowing domain-specific systems to potentially benefit from advances in general-purpose MLLMs without repeatedly adapting the language model. Future work will explore richer structured forensic context and evaluate the Perception-as-Tool paradigm across different general-purpose MLLMs and domain-specific multimodal tasks.


\section*{Acknowledgments}
This work was supported  by the National Natural Science Foundation of China (Grant Nos. 62376070 and 62076195), the Fundamental Research Funds for the Central Universities (AUGA5710028726), and the China Postdoctoral Science Foundation (Grant No. 2026M794773).

\bibliographystyle{named}
\bibliography{ijcai26}

\end{document}